\documentclass[10pt,A4]{article}

\usepackage{ioa_style}

\usepackage{tabularx}
\usepackage{array}
\usepackage{ragged2e}
\usepackage{graphicx}
\usepackage{amsmath,amsfonts,bm}
\usepackage{courier}
\usepackage[colorlinks=true,urlcolor=blue,linkcolor=black,citecolor=black]{hyperref}
\usepackage{comment}

\begin{document}
\title{\RaggedRight{\uppercase{\bf{On Vision Transformers for Classification Tasks in Side-Scan Sonar Imagery}}}}

\author{
	\begin{tabularx}{\textwidth}{@{}>{\RaggedRight}p{28mm} @{\;}>{\RaggedRight}X}
 	{BW Sheffield} & {NSWC PCD, Panama City, Florida, USA} \\
        {Jeffrey Ellen} & {NIWC, San Diego, California, USA} \\
        {Ben Whitmore} & {NIWC, San Diego, California, USA} \\
	\end{tabularx}
}

\date{} 
\maketitle
\thispagestyle{fancy}

\section{Introduction}

Traditionally, the classification of SSS imagery has relied heavily on manual interpretation by experts, supplemented by conventional machine learning techniques that utilize hand-crafted features. These methods, while effective to a degree, are time-consuming and often fall short in capturing the complex, varied textures and structures present in underwater environments. The advent of Convolutional Neural Networks (CNNs) marked a significant advancement in this field, offering more robust feature extraction capabilities and automating the classification process to a large extent.

\begin{figure}[!htb]
\centering
\includegraphics[width=0.4\textwidth]{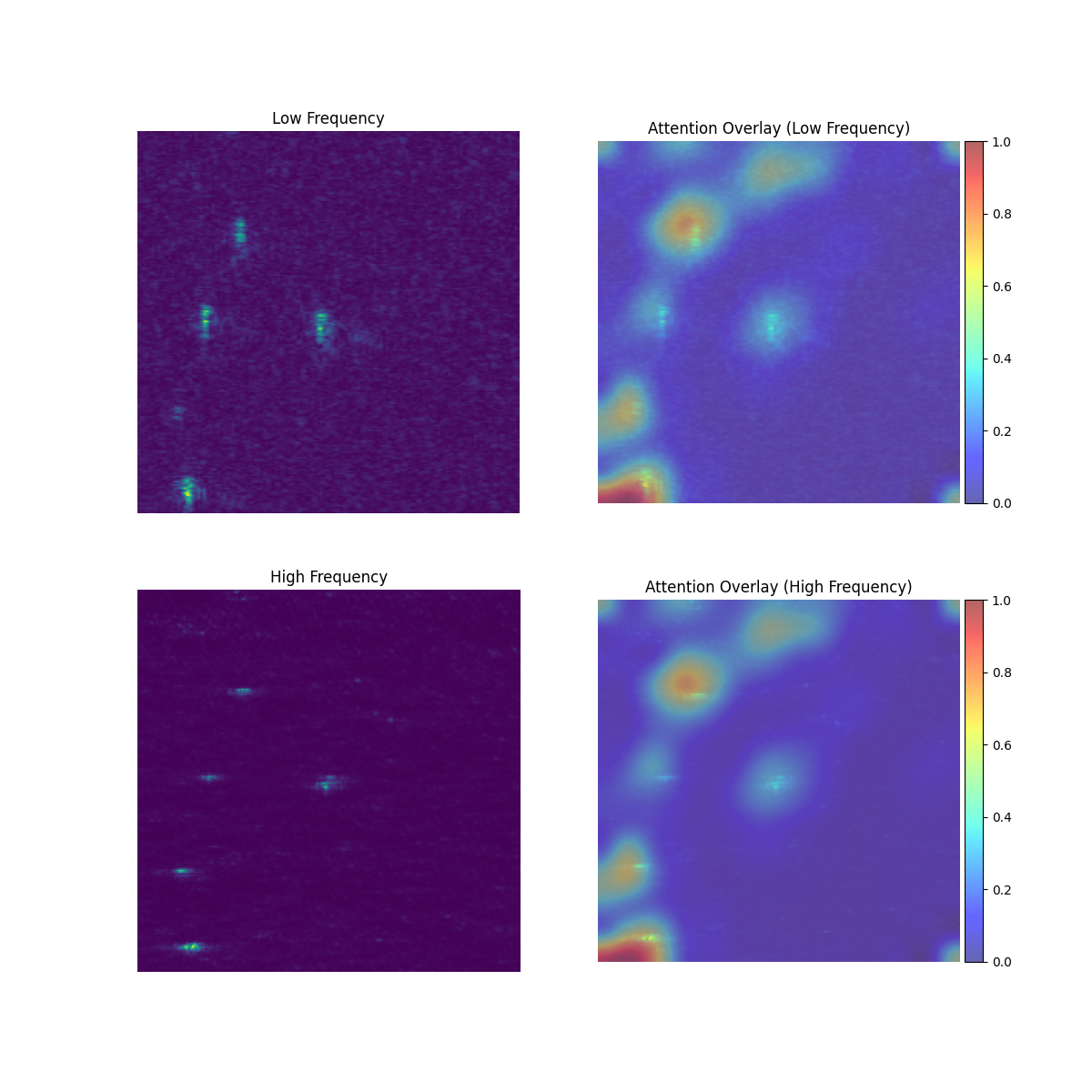}
\caption{Low and high frequency channels of synthetic aperture sonar imagery show what the Vision Transformer pays the most attention to on the seafloor, rocks, when determining if a man-made object exists.}
\label{fig:attention}
\end{figure}

However, the introduction of Vision Transformers (ViTs) has opened new avenues for SSS imagery analysis. ViTs leverage the power of self-attention mechanisms to process images as sequences of patches, allowing for a more flexible and comprehensive understanding of the spatial hierarchies within the data. This approach holds promise for capturing the intricate details and global context of SSS images, potentially surpassing the performance of traditional CNNs in classifying complex underwater scenes. Unlike CNNs, which primarily focus on local features due to their convolutional nature, ViTs can process entire image patches at once, allowing them to better understand the broader context of a scene. The self-attention mechanism in ViTs may be of benefit for particular classification scenarios where seafloor bottom types such as rocky and ripple sand negatively impact CNNs in reporting false alarms when searching for man-made objects.

The objective of this paper is to rigorously compare the efficacy of ViTs with established CNN models for binary image classification tasks in SSS imagery. By conducting a comprehensive evaluation based on a variety of metrics, as well as considerations of computational efficiency, we aim to provide a detailed empirical assessment of each model architecture's strengths and limitations to benefit future sonar machine learning research.

\section{Related Work}

Vision Transformers have shown excellent results compared to state-of-the-art convolutional networks while requiring substantially fewer computational resources to train \cite{dosovitskiy2020image}. The main contribution of Dosovitskiy et al. \cite{dosovitskiy2020image} was to adapt the concept of attention, common in natural language processing applications to images. They were not the first to try, however, instead of computing attention between every pairwise set of pixels as predecessors had done, Dosovitskiy et al. \cite{dosovitskiy2020image} proposed computing the metric at two distinct spatial scales; first in complete 16x16 local blocks, then computing the metric between individual blocks at the scale of the whole image. This technique exceeded existing benchmarks on natural images from ImageNet, CIFAR, and on subsets of Pets and Flowers \cite{dosovitskiy2020image}.

ViTs have also been shown to have superior classification performance in remote sensing images \cite{zhang2021integrating}, which are more similar to SSS images in that they are downward looking and are absent in most deep learning image libraries. Multiple surveys in ViTs have been conducted recently, including a survey on transformers in remote sensing \cite{deininger2022comparative, mauricio2023comparing,  aleissaee2023transformers}. Applications of ViTs in image classification tasks have exploited additional metadata sources using a multi-modal approach in medical imaging \cite{pacheco2021attention}, remote sensing \cite{roy2023multimodal, yao2023extended}, and in SSS \cite{ellen2023improving}.

 From the domain of underwater acoustics in automated target recognition tasks, a study\cite{li2022stm} introduces the Spectrogram Transformer Model (STM), a novel application of the Transformer architecture for recognizing underwater acoustic targets exhibiting superior performance over traditional CNNs. A different paper presents the Dual-Path Vision Transformer Network (DP-ViT) \cite{sun2022dp}, designed for accurate target detection in sonar imagery, including both forward-look and side-scan sonar.

\section{Methodology}

\subsection{Model Architectures}

In this study, multiple architectures are empirically evaluated against one another, namely two CNN architectures and two ViT architectures: ResNet\cite{he2016deep}, ConvNext\cite{liu2022convnet}, ViT\cite{dosovitskiy2020image}, and SwinViT\cite{liu2021swin}. Model sizes that are considered tiny(T), small(S), and base(B) are chosen, such as ViT-T or ConvNext-S, ignoring larger variants like large and huge. Every model was configured to operate on dual channel, low and high frequency, SSS imagery with 224x224 snippet sizes along with the last fully-connected layer changed to the number of target classes.

\subsection{Dataset}

Data was collected using a dual frequency synthetic aperture sonar (SAS) from several different geographical locations with varying seafloor bottom types. Due to the nature of SAS producing large resolution imagery, snippets were extracted and resized to a common 224x224 size that either contained a man-made object or not. All datasets are perfectly balanced with positive and negative instances.

\begin{table}[h]
\centering
\begin{tabular}{|c|c|c}
\hline
\textbf{Dataset} & \textbf{Number of Samples} \\ \hline
Train & 6878 \\ \hline
Validation & 1204 \\ \hline
Test & 1244 \\ \hline
\end{tabular}
\caption{Number of samples for each dataset; perfectly balanced with positive and negative instances.}
\label{tab:dataset_statistics}
\end{table}

\subsection{Training}

Each model is trained using PyTorch on a Nvidia A6000 graphics processing unit. Pretrained weights on ImageNet-1K were used to initialize each deep neural network. To reduce overfitting, each model is optimized against the loss produced from the validation dataset to alleviate overfitting to the training dataset.

Various data augmentations are used during training such as horizontal flip, +- 5 degree rotation, subtle affine, and zoom. No data augmentations are applied on the validation and test dataset as the objective is to evaluate on real imagery. Careful consideration has to be placed on what augmentations make the most sense for sonar imagery. For example, vertical flipping essentially reverses the direction of the sonar's line of sight, which is not a realistic representation of how SAS data is acquired.

\begin{table}[h]
\centering
\begin{tabular}{|c|c|}
\hline
\textbf{Hyperparameters} & \textbf{Values}  \\ \hline
Channels                & 2     \\ \hline
Batch Size              & 32          \\ \hline
Optimizer               & AdamW   \\ \hline
Loss                    & Binary Cross Entropy      \\ \hline
Learning Rate           & 0.00008      \\ \hline
Patience                & 20  \\ \hline
\end{tabular}
\caption{Hyperparameters used for training the deep learning models. The number of channels represents the low and high frequency input channels. The learning rate controls the step size at which the model's weights are updated, and patience refers to the number of epochs with no improvement in validation loss before early stopping is triggered.}
\label{tab:hyperparameter_table}
\end{table}

\section{Evaluation Metrics}

\subsection{Classification Performance}

The effectiveness of the models are measured with four widely applied evaluation criteria: f1-score, recall, precision, and accuracy. These metrics provide a comprehensive evaluation of the model's performance, quantifying its ability to accurately identify true positives(TP) and its capacity to limit false positives(FP). Depending on the scenario, the cost of classifying a scene as a false negative(FN) or simply not classifying a scene correctly when it is actually a true negative (TN) may be quite costly.

$$\text{Precision} = \frac{\text{TP}}{\text{TP} + \text{FP}}$$

Precision measures the models ability to return only relevant instances, calculating the fraction of instances correctly identified as objects out of all instances that the model classified as such.

$$\text{Recall} = \frac{\text{TP}}{\text{TP} + \text{FN}}$$

Recall measures the model's ability to identify all relevant instances, calculating the fraction of actual objects that the model correctly identified.

$$\text{F1-score} = 2 \times \frac{\text{Precision} \times \text{Recall}}{\text{Precision} + \text{Recall}}$$

The F1-score is the harmonic mean of Precision and Recall, providing a balanced representation of these two metrics.

$$\text{Accuracy} = \frac{\text{TP} + \text{TN}}{\text{TP + TN + FP + FN}}$$

Accuracy measures the overall effectiveness of the model in capturing both positive and negative instances. However, in the case of SSS imagery, accuracy can be misleading in imbalanced scenarios where man-made objects are rare compared to the abundant background clutter such as rocks. Since all the datasets are balanced, this is not an issue.

\subsection{Computational Efficiency}

Inference time is measured by averaging each model's inference across $n$ runs. To ensure accurate measurements, inference time is only recorded when the data is loaded on the device, and the GPU warm-up process has been completed. The GPU warm-up involves running a few inference iterations to allow the GPU to reach a stable state and optimize its performance. Time is measured asynchronously, capturing the duration from the start to the end of the GPU computation using CUDA events.

$$\text{Average Inference Time} = \frac{\sum_{i=1}^{n} \text{Inference Time}_i}{n}$$

Throughput measures the maximum amount of parallelization, indicating the number of instances that can be processed per second. It is calculated as:

$$\text{Throughput} = \frac{\text{Number of batches} \times \text{Batch Size}}{\text{Total time}}$$

Floating Point Operations per Second (FLOPS) quantifies the number of floating-point operations (addition, subtraction, multiplication, and division) performed in one second. This unit of measure is used to assess the computational cost or efficiency of the ViT. FLOPS are calculated using the python library \textit{fvcore} by Meta AI Research.

$$\text{FLOPS} = \frac{N_a + N_s + N_m + N_d}{T}$$

The number of parameters, or weights, in a model is closely related to its capacity to learn. A larger number of weights generally requires a larger dataset for training and thus results in a larger model size. The model size has implications for memory requirements and the feasibility of deploying the ViT in resource-constrained environments. Only the learnable parameters, ones that require gradients, are counted.

$$\text{Total Learnable Parameters} = \sum_{p_i \in P} p_i$$

\section{Discussion of Results}

\begin{table}[!htb]
\centering
\begin{tabular}{lcccc}
\hline
Model & F1-Score (\%) & Recall (\%) & Accuracy (\%) & Precision (\%) \\ \hline
ViT-T & 99.12 & 98.67 & 99.16 & 98.62 \\
ViT-S & 99.36 & \textbf{100.00} & 99.16 & 97.80 \\
ViT-B & \textbf{100.00} & 99.80 & \textbf{100.00} & 99.16 \\
Swin-T & 98.23 & 96.38 & 98.40 & 99.31 \\
Swin-S & 99.06 & 97.24 & 98.99 & \textbf{100.00} \\
ResNet-18 & 98.84 & 98.91 & 98.82 & 97.92 \\
ResNet-50 & 98.64 & 98.14 & 98.49 & 98.21 \\
ResNet-101 & 99.00 & 97.51 & 98.91 & 99.56 \\
ConvNext-T & 98.29 & 97.50 & 98.49 & 98.48 \\
ConvNext-S & 99.07 & 99.23 & 99.07 & 97.99 \\
\hline
\end{tabular}
\caption{Relative performance metrics for the various ViT and CNN models where 100\% represents the best performance in it's category.}
\label{tab:model_relative_performance}
\end{table}

\begin{table}[h]
\centering
\begin{tabular}{lccccc}
\hline
Model & \begin{tabular}[c]{@{}c@{}}Inference Speed\\ (ms)\end{tabular} & \begin{tabular}[c]{@{}c@{}}Throughput\\ (imgs/sec)\end{tabular} & \begin{tabular}[c]{@{}c@{}}FLOPS\\ (G)\end{tabular} & \begin{tabular}[c]{@{}c@{}}Parameters\\ (M)\end{tabular} & \begin{tabular}[c]{@{}c@{}}Model\\ Size (MB)\end{tabular} \\ \hline
ViT-T & 8.22 & 3890.37 & \textbf{34.23} & \textbf{5.48} & \textbf{20.89} \\
ViT-S & 18.01 & 1816.96 & 135.39 & 21.57 & 82.27 \\
ViT-B & 48.04 & 664.50 & 538.50 & 85.60 & 326.55 \\
Swin-T & 28.00 & 1141.62 & 144.12 & 27.52 & 104.98 \\
Swin-S & 48.52 & 670.56 & 280.41 & 48.84 & 186.30 \\
ResNet-18 & \textbf{7.04} & \textbf{4578.46} & 56.93 & 11.17 & 42.63 \\
ResNet-50 & 21.56 & 1483.78 & 130.24 & 23.51 & 89.67 \\
ResNet-101 & 37.97 & 821.54 & 249.36 & 42.50 & 162.12 \\
ConvNext-T & 20.81 & 1561.18 & 142.88 & 27.82 & 106.12 \\
ConvNext-S & 34.64 & 922.17 & 278.39 & 49.45 & 188.65 \\
\hline
\end{tabular}
\caption{The computational efficiency of each model is measured in terms of average inference speed, throughput, FLOPS, number of parameters, and model size which are critical for real-time applications especially in resource-constrained environments such as underwater vehicles.}
\label{tab:model_metrics}
\end{table}

ViTs performed the best compared to popular CNN architectures in terms of binary image classification performance as seen in Table \ref{tab:model_relative_performance}. Despite having better classification performance, deployment of such models often fall under constraint of memory especially on edge devices located in underwater vehicles. As a reminder, this study did not include other ViT variants such as large and huge which are impractical for real-time inference in modern underwater vehicles. 

ViTs in general require more computational resources due to their higher parameter count. ResNet-101 is considered fairly steep in modern deep learning CNNs however when compared to ViT-B in Table \ref{tab:model_metrics}, it has less than half the parameter count. It should also be noted that inference speed is generally slower for ViTs as well. Potential future directions would dive deeper into model optimization techniques such as quantization and pruning to further examine classification performance over computational efficiency trade-offs.

It is widely accepted that ViTs require more training data than CNNs to obtain good predictive performance. The requirement of larger training data for ViTs distill down to what are called inductive biases which encode assumptions or prior knowledge into the design of each machine learning algorithm. Inductive biases in CNNs are local connectivity, weight sharing, hierarchical processing, and spatial invariance. These inductive biases make machine learning algorithms computationally more feasible and/or exploit domain knowledge. Because CNNs have more inductive biases than ViTs, less training data is required to learn patterns or features in an input image.

\section{Conclusion}

Final results revealed that ViT-based models outperformed classification performance over popular CNN models in diverse environmental settings captured with SSS. However, ViTs do have computational efficiency trade-offs that prohibit them from real-time deployment applications especially in limited resource environments such as underwater vehicles. CNNs still have a strong role to play in the practical deployment of image classification tasks using SSS due to their built-in inductive biases.

Future ViT research goals in SSS will examine the benefits of self-supervised learning with unlabeled data and multi-modal applications where additional underwater vehicle data is fused in order to inform predictive performance.

\bibliographystyle{unsrt} 
\bibliography{sources}

\end{document}